\documentclass[11pt]{article}

\usepackage{graphicx}
\graphicspath{{50Images/}}
\usepackage{multirow}
\usepackage{url}
\usepackage{soul}
\usepackage{amsmath}
\usepackage{color}
\usepackage[normalem]{ulem}

\usepackage{paclic33}
\usepackage{times}
\usepackage{latexsym}
\usepackage{amsmath}
\usepackage{multirow}
\usepackage{url}

\title{Identifying Adversarial Sentences by Analyzing Text Complexity}

\author{Hoang-Quoc Nguyen-Son$^\textsuperscript{1}$, Tran Phuong Thao$^\textsuperscript{2}$, Seira Hidano$^\textsuperscript{1}$, and Shinsaku Kiyomoto$^\textsuperscript{1}$ \\
  $^\textsuperscript{1}$KDDI Research, Inc. \\
  2-1-15, Ohara, Fujimino, Saitama, 356-8502, Japan\\
  {\tt \{ho-nguyen, se-hidano, kiyomoto\}@kddi-research.jp} 
  \\
  $^\textsuperscript{2}$The University of Tokyo \\
  7-3-1, Hongo, Bunkyo, Tokyo, 113-8656, Japan\\
  {\tt tpthao@yamagula.ic.i.u-tokyo.ac.jp} 
  }

\date{}

\begin{document}
\maketitle
\begin{abstract}

Attackers create adversarial text to deceive both human perception and the current AI systems to perform malicious purposes such as spam product reviews and fake political posts. 
We investigate the difference between the adversarial and the original text to prevent the risk.
We prove that the text written by a human is more coherent and fluent.
Moreover, the human can express the idea through the flexible text with modern words while a machine focuses on optimizing the generated text by the simple and common words.
We also suggest a method to identify the adversarial text by extracting the features related to our findings.
The proposed method achieves high performance with 82.0\% of accuracy and 18.4\% of equal error rate, which is better than the existing methods whose the best accuracy is 77.0\% corresponding to the error rate 22.8\%.

\end{abstract}

\section{Introduction}

The computer-generated text has achieved remarkable success in replacing human roles in interactive systems such as question answering and machine translation.
However, aside the positive impacts, an adversary takes advantage of the text to fool the judgment systems which are even unrecognized by human-beings themselves.
The fake political attitudes and product previews, for instances, have significantly affected the awareness of real audiences. 
It raises the urgent task which has to efficiently identify the adversarial text before it is spread through the public media.

Previous methods have focused on proposing classifications to detect computer-generated text that is used in various unscrupulous purposes.
More particularly, \newcite{labbe2013duplicate} created a detector\footnote{\url{http://scigendetection.imag.fr/main.php}} to identify a dummy paper by using text similarity.
Other methods recognized untrusted information from machine translation text based on $N$-gram model~\cite{aharoni2014automatic} and word matching~\cite{nguyen2018identifying,nguyen2019detecting}.
However, according to the dramatic development of enhanced technologies, especially in deep learning era, adversarial text generated from the deep networks~\cite{iyyer2018adversarial,liang2018deep} can bypass in both existing work and human recognition.

\begin{figure}[t]
\centering
\includegraphics[]{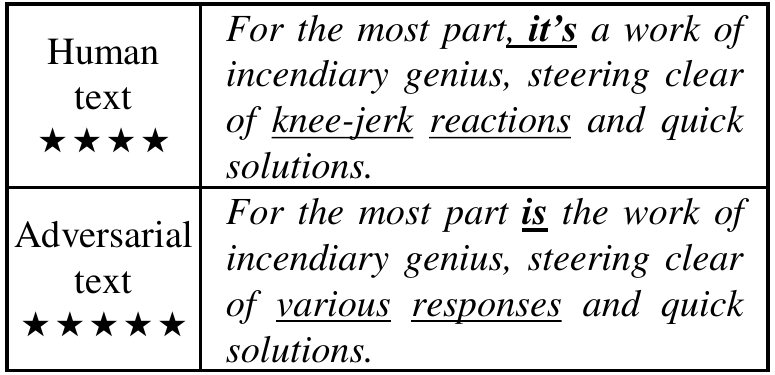}
\caption{Human-created original vs machine-generated adversarial text.}
\label{Fig_1_Example}
\end{figure}

We aim to investigate the adversarial texts~\cite{iyyer2018adversarial} which have different syntax structure with the original text due to the difficulty in tracing their origins.
Such texts are more dangerous than other easily tractable texts which are simply created by word or phrase modification~\cite{liang2018deep,ebrahimi2018hotflip}.
Moreover, the adversarial texts considered in this paper really fool AI systems.
One of the actual adversarial texts is picked from the development set as represented in Figure~\ref{Fig_1_Example}.
The samples are movie reviews whose interest rates are represented with the number of stars determined by a common sentiment analysis system~\cite{socher2013recursive}.
In these reviews, the level of the adversarial text is labeled with four-star, the original is more interested with five.

The original sentence is often more complex than the adversarial text in both word usage and text structure.
Human writers express their intentions via fashionable and modern words, such as ``\textit{\underline{knee-jerk}}'' and ``\textit{\underline{reactions}}''. 
In contrast, the adversarial text is optimized its readability by simple common words, i.e. ``\textit{\underline{various}}'' and ``\textit{\underline{responses}}.''
Moreover, human tends to use flexible structural utterances. 
The flexibility is illustrated with the use of the complex expression ``\textit{most part\underline{\textbf{, it's}} a work}'' instead of ``\textit{most part \underline{\textbf{is}} the work}.''

\paragraph{Contribution} 
In this paper, we analyze the most threatening adversarial text which not only fools the recent AI system but also is difficultly tracked because of changing the original structure.
We thereafter propose the following process to distinguish the adversarial with the original text:

\begin{itemize}
\item
We estimate the text coherence by matching words and measuring the word similarities.
Only the high similarities which mainly construct the coherence are distributed into certain groups depending on the part of speech (POS) of the matched words.
In each group, the similarities are normalized by the means and the variances that represent for the coherence features.  
\item
We suggest using frequencies to exploit the difference of word usage in original and adversarial text. 
In particular, the frequencies are allocated to appropriate POS groups and are used as frequency features.
\item
We design other features to address the optimization problems of generating adversarial text. 
More especially, we notice that the adversarial text is short and may contain successive duplicate phrases.
We thus integrate sentence length with the number of duplicate phrases in order to extract the optimization features.
\item
We combine our features with the features extracted from the $N$-gram language model for determining whether the input sentence is an original text written by a human or an adversarial text generated by a machine.
\end{itemize}

To evaluate our method, we use 11K original sentences from a common movie review corpus\footnote{\url{http://nlp.stanford.edu/~socherr/stanfordSentimentTreebank.zip}}. 
We then generate the corresponding adversarial text using the syntax-based system by \newcite{iyyer2018adversarial}.
Afterward, we select approximate 1500 pairs which are classified in different sentiment labels by a Stanford system~\cite{socher2013recursive}.
The result shows that our method achieves high performance with 82.0\% of the accuracy and 18.4\% of the equal error rate.
It outperforms the existing state-of-the-art method whose accuracy is 77.0\% and equal error rate is 22.8\%.

\paragraph{Roadmap}
The rest of this paper is organized as follows. The related work is presented in Section~\ref{section:related_work}. 
Our proposed method is described in Section~\ref{section:proposed_method}. 
The evaluation is given in Section~\ref{section:evaluation}. 
Finally, Section~\ref{section:conclusion} summarizes some main key points and mentions future work.

\section{Related Work}
\label{section:related_work}
In this section, we present previous work in two aspects: methods for generating adversarial texts and methods for identifying adversarial texts.

\subsection{Adversarial Text Generation}

There are two approaches for generating adversarial text: non-syntax-based and syntax-based. 
In the first approach, an adversary modifies some parts of the original text but still preserve its structure.
On the other hand, in the second approach, the adversary changes the text's syntax to deceive the AI systems.

\paragraph{Non-syntax-based Approach}

\newcite{liang2018deep} changed the salient text components via white-box attack using cost gradients or black-box attack using occluded samples. 
The modification can apply for both \emph{character} and \emph{word} levels in order to generate robust adversarial text which efficiently fools a multiclass classifier related to news posts.
Also targeting on this classifier, \newcite{ebrahimi2018hotflip} generated the adversarial text by using a set of operations on \emph{characters} such as flip, insertion, and deletion based on the one-hot vectors extracted from the input text.
Besides the multiclass classification, the adversarial text is also able to attack the question answering (QA) system. 
For instance, \newcite{jia2017adversarial} padded a distract \emph{sentence} into text for changing the answers of the QA system.
Furthermore, \newcite{ribeiro2018semantically} can generate the adversarial text which deceives both AI systems including QA and sentiment analysis.
More especially, the authors used a set of rules on \emph{phrase} to generate adversarial questions which look alike to the origins but change the results of these systems.
\newcite{alzantot2018generating} also proved that the adversarial text affords to fool various AI systems, namely sentiment analysis and textual entailment, using \emph{word} replacements. 

\paragraph{Syntax-based Approach}
Most previous work from the non-syntax-based approach that we mentioned above adapts the operations such as modifications, insertions, or deletions on various text levels: characters, words, phrases, and sentences. 
Due to the unchanged structures, such texts easily trace back to their origins and these generated texts are easily filtered.
In the opposite way, the syntax-based approach addresses more serious adversarial texts when the structures are changed, so such texts are easy to mix with their origins without being detected.
We, therefore, focus on this approach instead of the other. 
In the syntax-based approach, \newcite{iyyer2018adversarial} generated a paraphrase with a desired syntax by using attention networks to transfer the text structure.
Such adversarial texts can target to two current popular risks: (i) fake reviews by fooling sentiment analysis system, and (ii) political posts by deceiving the textual entailment.

\subsection{Adversarial Text Detection}

Previous work is categorized into four approaches: parse tree, word distribution, $N$-gram model, and word similarity.

\paragraph{Parse Tree}
\newcite{li2015machine} prove that the syntactic structure of a human-written sentence is more complex than that of computer-generated one because the simple artificial text is often created to prevent the mistakes in both grammar and semantic.
The structure of the simple text is well-balanced, so the authors extracted some related features, i.e. the ratio of right/left-branching notes in various scopes: main constituents and whole sentence.
Some surface and statistical features were also used to including parse tree depth, sentence length, and out-of-vocabulary words.
The main drawback of parse tree approach is that it only investigates on text syntax but ignore the semantics itself.

\paragraph{Word Distribution}
The word distribution in the large text is used to classify computer- and human-generated text.
For example, \newcite{labbe2013duplicate} indicate the high similarity of the distribution in artificial documents.
They suggested a metric, namely inter-textural distance, to measure the similarity between two texts.
It can be used to identify fake academic papers with impressive accuracy.
More general, \newcite{nguyen2017identifying} used Zipfian distribution to identify other texts.
Additional features extracted from humanity phrases (e.g., idiom, clich{\'e}, ancient, dialect, and phrasal verb) and co-reference resolution were also applied to improve their result.
The main drawback of word distribution approach is that a large number of words are required.
This limitation is also confirmed by the authors of both the inter-textual distance and the Zipfian distribution.

\paragraph{$N$-gram Model}

The common method to estimate the fluency of continuous words is to use the $N$-gram model.
Many researchers have measured this property on discontinuous words and combine with the $N$-gram model.
For instance, \newcite{arase2013machine} used sequential pattern mining to extract fluent human patterns such as ``\textit{not only * but also}'' and ``\textit{more * than}.'' They contrasted with the weird patterns (e.g., ``\textit{after * after the}'' and ``\textit{and also * and}'') in machine-generated texts from low-resource languages.
\newcite{nguyen2017detecting} extracted features from two types of noise words: (i) the humanity words from a user message, such as misspelled (e.g., comin, hapy) and short-form/slang words (e.g., tmr, 2day), (ii) the untranslated words from a machine message.
This approach, however, is only suitable for social network texts that abundantly contained substantial noise words.
On the other hand, \newcite{aharoni2014automatic} targeted functional words that are often chosen by a machine for improving the readability of the generated text. 

\paragraph{Word Similarity}
\newcite{nguyen2018identifying} proposed the classification based on the idea that: the coherence between words in a computer-generated text is less than that in a human-generated text. 
They matched similar words in every pair of sentences in a paragraph using Hungarian maximum matching algorithm.
More particularly, each word was matched with the most similar word in another sentence. 
The drawback of this work is that the relationships between words inside a sentence are not considered so it cannot be applied for individual sentences as targeted in this paper.
\newcite{nguyen2019detecting} overcome the limitation by matching similar words in the whole text. 
The maximum similarity for each word was used to estimate the coherence while the other similarities are dismissed.
For coherence features mentioned in the Section~\ref{section:Matching_similar_words}, we indicate that the other high similarities are also useful to measure the text coherence and efficient to identify the adversarial text.

\section{Proposed Method}
\label{section:proposed_method}

The overview of the proposed method is formalized as four parallelizable steps:
\begin{itemize}
\item
\textbf{Step 1 (\textit{Matching words})}: 
Each word is matched with the other words, and their similarities are estimated by Euclidean distances in word embedding.
These similarities represent the connections of words and are thus used to extract coherence features.
\item
\textbf{Step 2 (\textit{Estimating frequencies})}: 
The frequencies of individual words are inferred from corresponding items of Web 1T 5-gram corpus. 
The frequency indicates the popularity of the word in various context usages.
\item
\textbf{Step 3 (\textit{Finding optimization issues})}:
Optimization issue features which result from the optimization process of adversarial text generation are extracted from sentence length and successive duplicated phrases.
\item
\textbf{Step 4 (\textit{Extracting word $N$-gram})}:
The text fluency is measured by using the word $N$-gram model. 
$N$ continuous words with $N$ between 1 to 3 are used for this model.
\end{itemize}

The details of each step using the examples mentioned in Figure~\ref{Fig_1_Example} are described in the following sub-sections.

\subsection{Matching Words (Step 1)}
\label{section:Matching_similar_words}

\begin{figure*}[t]
\centering
\includegraphics[]{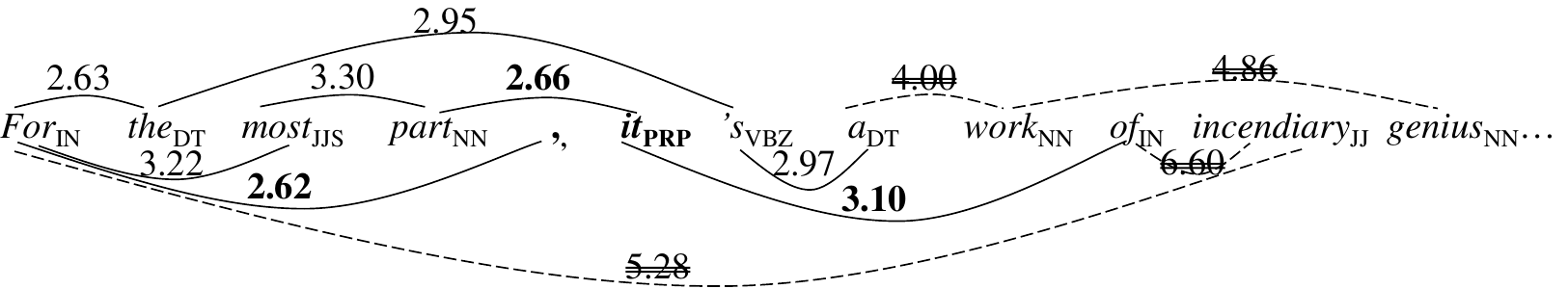}
\caption{Matching words in the human original text.}
\label{Fig_3_Matching_similar_words_human}
\end{figure*}

Words in the input text are separated and tagged with the part-of-speeches (POSs) using a Stanford tagger~\cite{manning2014stanford}.
Each word is matched with the other words, then their similarities are estimated.
For inferring the similarity between two words, we measure the Euclidean distance of their vectors in word embeddings. 
The higher similarity of two words results in the lower distance.
The GloVe corpus~\cite{pennington2014glove} is used to map the words, collected from Wikipedia and Gigaword, with 300-dimensional vectors. 
The similarities of some words in the human-generated text are illustrated in Figure~\ref{Fig_3_Matching_similar_words_human} with the POS taggers denoted by subscripts.

\begin{figure*}[t]
\centering
\includegraphics[]{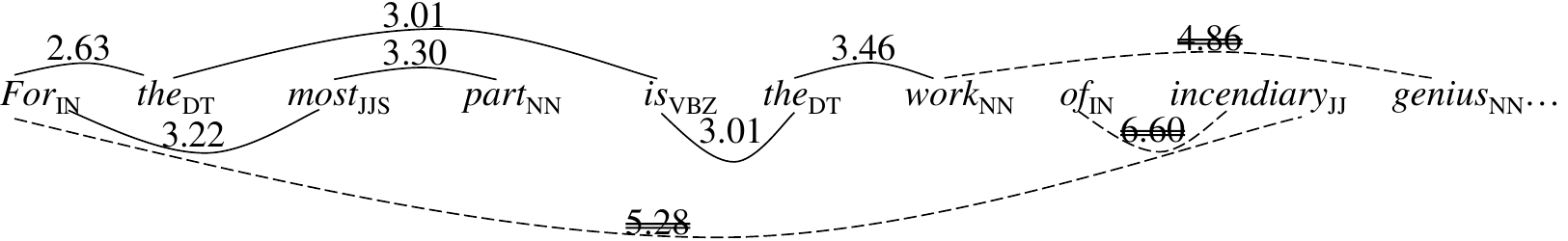}
\caption{Matching words in the machine adversarial text.}
\label{Fig_4_Matching_similar_words_machine}
\end{figure*}

A machine tends to create a simple text so that the text's meaning and readability are preserved.
The generated text is thus generally shorter than the human-generated one, as also claimed by \newcite{volansky2013features}.
The long expression of the original text compared to the short of the adversarial text is shown in Figure~\ref{Fig_3_Matching_similar_words_human} and Figure~\ref{Fig_4_Matching_similar_words_machine}, respectively.
The additional words and their connections with other words are marked in bold to emphasize the difference.
The small values of the distances demonstrate the tight connections that are created by these padding words.
These connections do not influence the overall meaning but slightly improve the text coherence.

The high similarities with small distances have higher impacts on the text coherence than the low similarities.
Therefore, we only choose the highs and eliminate the others. 
We suggest a threshold of $\alpha$ to determine the ratio between them. 
The $\alpha$ is set to 0.7 after being optimized from the development set as mentioned in Section~\ref{section:evaluation}.
It means that only 70\% of the high similarities are selected while the remaining is removed as presented by double strike-though numbers linked to dashed-line connections.

\begin{figure*}[t]
\centering
\includegraphics[]{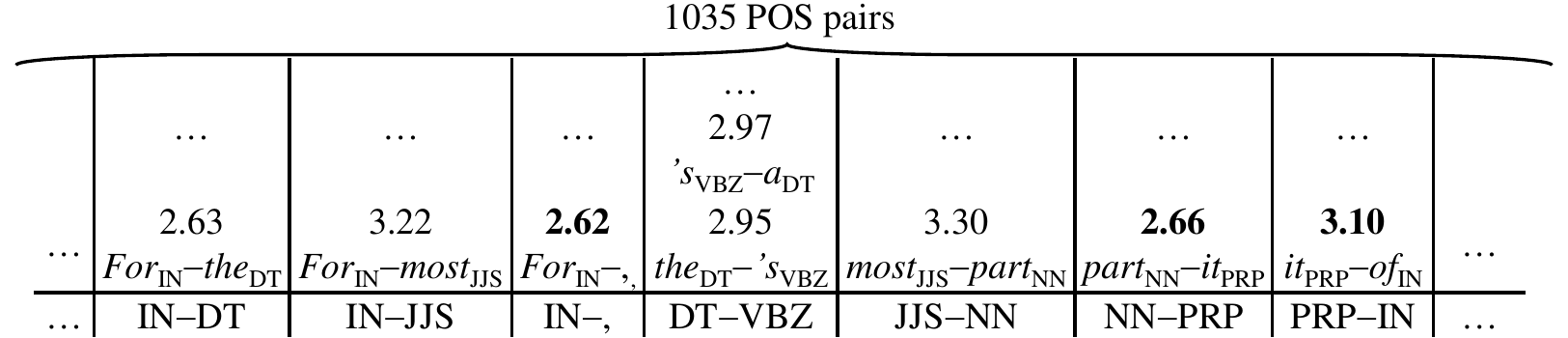}
\caption{Distributing high similarities of the human text.}
\label{Fig_5_Distributing_similarities}
\end{figure*}

According to the POS tags, words play different roles in a certain sentence.
Major words like nouns (NN) and verbs (VB) have higher influential than minor ones such as determiners (DT) and prepositions (IN).
We thus distribute the high similarities to appropriate POS groups.
In Figure~\ref{Fig_5_Distributing_similarities}, two pairs ``$\textit{the}_\text{DT}\text{--}\textit{'s}_\text{VBZ}$'' (2.95) and  ``$\textit{'s}_\text{VBZ}\text{--}\textit{a}_\text{DT}$''(2.97) are allocated to the same group, i.e. DT\text{--}VBZ.
We use all of 45 POS tags containing in the training set and produce 1035 possible combinations in total.

The individual POS groups often contain different numbers of similarities.
The numbers in each group are normalized by using the means and the variances.
Total 2070 statistical values are calculated for all POS groups.
These values are used as the features representing the coherence of the text.

\subsection{Estimating Frequencies (Step 2)}

\begin{figure*}[t]
\centering
\includegraphics[]{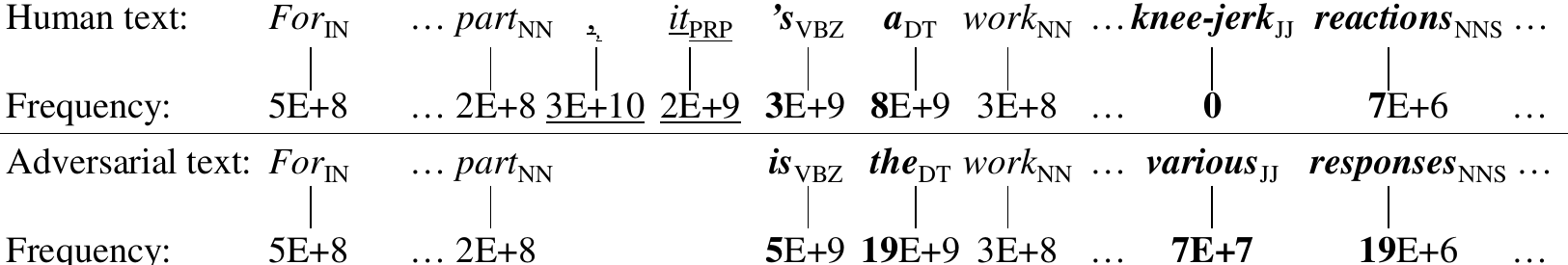}
\caption{Calculating word frequencies in the human and the adversarial text.}
\label{Fig_6_Calculating_frequencies}
\end{figure*}

After splitting and tagging POSs, we estimate the popularity of the words by using their frequencies in Web 1T 5-gram corpus\footnote{\url{https://catalog.ldc.upenn.edu/LDC2006T13}}.
This corpus counts the number occurrences of around 14 million common words in approximately 95 billion sentences extracted from available web pages.  
The frequencies of several words in the human and the adversarial text are shown in Figure~\ref{Fig_6_Calculating_frequencies}.
The words occurring in the only human text are underlined while the other differences are marked in bold.
All non-displayed words are identical in the two texts.

The machine-generated text is often optimized with ``safe words'' which are commonly used in other contexts.
It explains that the frequencies of the adversarial words are slightly higher than those of human words.
More especially, among the synonyms, the adversarial text tends to select high-frequency words, for instance, ``\textit{responses}'' (19E+6) instead of ``\textit{reactions}'' (7E+6).
On the other hand, in case of the same word meaning in the context, the standard words such as ``\textit{is}'' and ``\textit{the}'' have higher selection's priorities than the words ``\textit{'s}'' and ``\textit{a},'' respectively.
The writing styles of native speakers are very flexible, they can creatively choose ``fashionable words'' fitting to the context. 
For example, since ``\textit{knee-jerk}'' is rarely used, it is out of the highest frequency words even in the large 1 terabyte tokens of the Web 1T 5-gram corpus.

\begin{figure*}[t]
\centering
\includegraphics[]{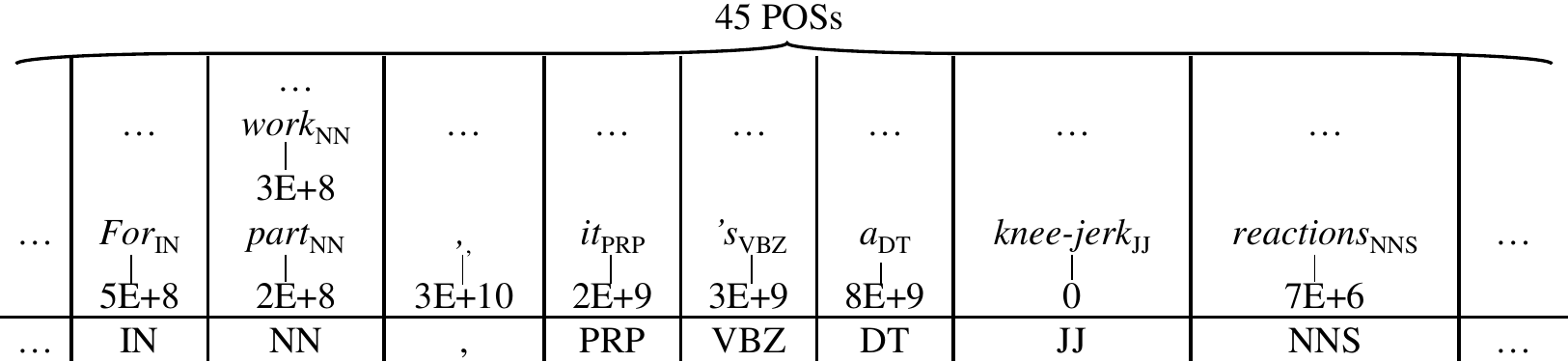}
\caption{Distributing word frequencies of the human text.}
\label{Fig_7_Distributing_frequencies}
\end{figure*}

Like the process of Step 1, we distribute the word frequencies into specific groups based on the POS tags. 
For instance, the two nouns ``\textit{part}'' and ``\textit{work}'' are delivered to the same group of nouns (NN) as illustrated in Figure~\ref{Fig_7_Distributing_frequencies}.
We also normalize the frequencies within the individual groups by the means and the variances for extracting the final frequency features.

\subsection{Finding Optimization Issues (Step 3)}

\begin{figure}[t]
\centering
\includegraphics[]{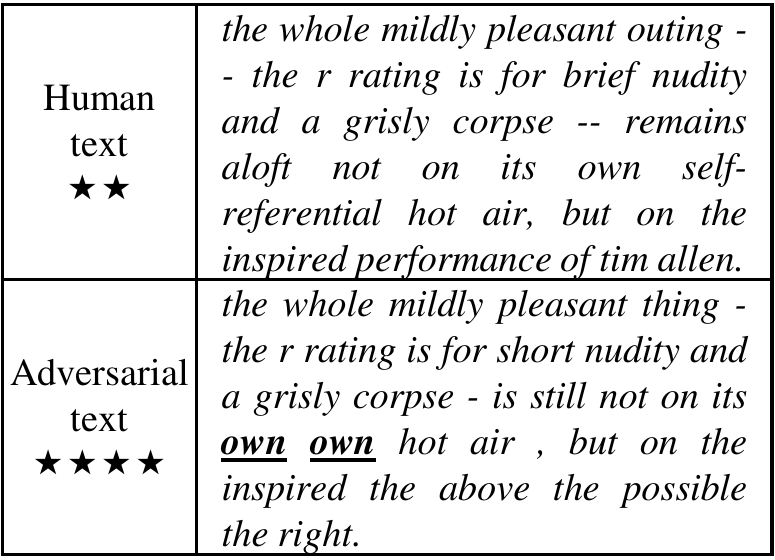}
\caption{Successive duplicated phrases in the adversarial text.}
\label{Fig_8_duplicated_phrases}
\end{figure}

The optimization process of adversarial text generation may cause the appearances of successive duplicated phrases. 
We, therefore, extracted the phrase-related features by counting the numbers of such phrases with the length varying from 1 to 5.
In Figure~\ref{Fig_8_duplicated_phrases}, since the adversarial sentence have two successive duplicate phrases ``\textbf{\underline{\textit{own}}},'' the 1-phrase-length feature is equal to 2.

Another issue of the optimization process is that a machine often generates a simple short text.
In other words, the machine practically selects candidates with a minimal number of words to express a certain intention.
Consequently, such generated texts are shorter than analogous texts written by a human.
To deal with this problem, we simply count the number of words and denote them as the length feature.
This length feature is integrated with the phrase-related features above and they are served as the optimization features.

\subsection{Extracting Word $N$-gram (Step 4)}

\begin{figure}[t]
\centering
\includegraphics[]{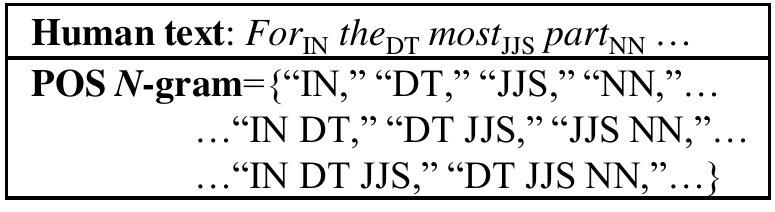}
\caption{Extracting POS $N$-gram from the human text.}
\label{Fig_9_POS_N_Gram}
\end{figure}

To evaluate the frequency of the text, we inherit the $N$-gram model to extract continuous POS phrases with the length up to 3.
Some extracted phrases from the human text are listed in Figure~\ref{Fig_9_POS_N_Gram}.
We use the POS tags for the model instead of the word because the similar phrases having the same structure can be recognized.
For example, the first pattern ``IN DT'' represents not only for the phrase ``\textit{For the}'' but also for other identical structural ones such as ``\textit{For a}'' and ``\textit{In the}.''

\section{Evaluation}
\label{section:evaluation}

\subsection{Dataset}

We created the experimental data by using 11,855 sentences from a movie review corpus\footnote{\url{http://nlp.stanford.edu/~socherr/stanfordSentimentTreebank.zip}}.
These sentences were inputted to the syntactically controlled paraphrase networks (SCPN)\footnote{\url{https://github.com/miyyer/scpn}} to generate adversarial text.
We only proceeded the input sentences which can produce the adversarial texts actually fooled the well-known sentiment analysis system~\cite{socher2013recursive}.
At the result, the 1489 inputs were considered as human-written text while the corresponding sentences were denoted as machine-generated text.

The 2978 satisfied sentences were split into three sets: for training, for development, and for test phases with the ratio as 60\%, 20\%, and 20\%, respectively.
To balance the human and the machine sentences in each set, we put a pair of the original and the adversarial sentences into the same set.
The development test was used to determine the threshold $\alpha$ described in Section~\ref{section:Matching_similar_words}.
Two pairs in development set are shown in Figure~\ref{Fig_1_Example} and Figure~\ref{Fig_8_duplicated_phrases}.

\subsection{Individual Features and Combinations}

We conducted experiments on our individual features and their combinations.
The experiments were run with three common machine learning algorithms including logistic regression (LOGISTIC), support vector machine (SVM) optimized by stochastic gradient descent (SGD(SVM)), and SVM optimized by sequential minimal optimization (SMO(SVM)).
The results are summarized in Table~\ref{tab:01Individual_and_combination} with individual features in the top rows and their combinations in the bottoms.
For assessing the performances on the test set, we used two standard metrics: the accuracy and the equal error rate (EER).

\begin{table*}[t]
\begin{center}
\begin{tabular}{c l c c c c c c}

\hline 
{\multirow{2}{*}{\textbf{No}}}&{\multirow{2}{*}{\textbf{Features}}} &\multicolumn{2}{c}{\textbf{LOGISTIC}} & \multicolumn{2}{ c }{\textbf{SGD(SVM)}} & \multicolumn{2}{c}{\textbf{SMO(SVM)}}  \\ 
&&\textbf{Accuracy} &\textbf{EER} &\textbf{Accuracy}&\textbf{EER} &\textbf{Accuracy}&\textbf{EER}\\ 
\hline
1 & Coherence features & 60.5\%    &    39.7\%    &    68.7\%    &    28.0\%    &    \textbf{73.8\%}    &    \textbf{25.6\%}  \\
2 & Frequency features &	\textbf{71.5\%}    &    \textbf{28.3\%}    &    69.0\%    &    28.4\%    &    69.2\%    &    30.0\%  \\

3 & Optimization features &	\textbf{70.2\%}    &    \textbf{29.8\%}    &    69.2\%    &    32.6\%    &    66.7\%    &    36.6\%  \\

4 & POS $N$-gram features &	57.8\%    &    41.7\%    &    65.4\%    &    35.1\%    &    \textbf{75.2\%}    &    \textbf{25.1\%}  \\

\hline 
5 & 1 + Frequency features &	60.8\%    &    38.3\%    &    72.5\%    &    \textbf{22.7\%}    &    \textbf{74.0\%}    &    26.6\%  \\

6 & 5 + Optimization features &	61.0\%    &    39.0\%    &    76.2\%    &    26.3\%    &    \textbf{77.7\%}    &    \textbf{23.2\%}  \\

7 & All features &	67.3\%    &    32.3\%    &    81.2\%    &    \textbf{\underline{14.7\%}}    &    \textbf{\underline{82.0\%}}    &    18.4\%  \\

\hline 

\end{tabular}
\caption{Individual features and combinations.}
\label{tab:01Individual_and_combination}
\end{center}
\end{table*}

\begin{table*}[t]
\begin{center}
\begin{tabular}{l c c c c c c}

\hline 
{\multirow{2}{*}{\textbf{Method}}} &\multicolumn{2}{c}{\textbf{LOGISTIC}} & \multicolumn{2}{ c }{\textbf{SGD(SVM)}} & \multicolumn{2}{c}{\textbf{SMO(SVM)}}  \\ 
&\textbf{Accuracy} &\textbf{EER} &\textbf{Accuracy}&\textbf{EER} &\textbf{Accuracy}&\textbf{EER}\\ 
\hline

\newcite{nguyen2017identifying} & 66.5\%    &    33.0\%    &    64.5\%    &    32.9\%    &    \textbf{67.3\%}    &    \textbf{25.9\%}  \\

\newcite{li2015machine} & 67.5\%    &    32.3\%    &    66.3\%    &    34.1\%    &    \textbf{68.7\%}    &    \textbf{31.1\%}  \\

\newcite{nguyen2019detecting} & 60.2\%    &    40.0\%    &    64.0\%    &    35.9\%    &    \textbf{73.3\%}    &    \textbf{21.1\%}  \\

\newcite{aharoni2014automatic} & 59.5\%    &    40.3\%    &    66.0\%    &    34.2\%    &    \textbf{77.0\%}    &    \textbf{22.8\%}  \\

\hline 
Our (All features) &	67.3\%    &    32.3\%    &    81.2\%    &    \textbf{\underline{14.7\%}}    &    \textbf{\underline{82.0\%}}    &    18.4\%  \\
\hline 

\end{tabular}
\caption{Comparison with other methods.}
\label{tab:02Comparison}
\end{center}
\end{table*}

In four groups of individual features, the experiment on optimization gives low results.
It indicates that the surface information extracted from the internal input sentence is insufficient to identify adversarial text.
The use of external knowledge such as the frequency can improve the performances.
However, the frequency is limited to separate words and ignore the mutual connections of them.
On the other hand, the coherence features based on these connections improve both accuracy and EER metrics.
The POS $N$-gram features achieve better performances and point out the low fluency of the adversarial text.

In combinations, while the frequency features target on individual words, the coherence features examine the mutual connections among them. 
They support each other to raise the overall performances.
The combination with the features based on optimization problems of adversarial generators even achieves better results.
Finally, each individual exploits the different aspects of adversarial problems, so all features put together can establish the new milestone with the best accuracy up to 82.0\%.

\subsection{Comparison}

We compare our method with previous work on identifying machine-generated text.
The results of the comparison are provided in Table~\ref{tab:02Comparison} with the highest performances marked in bold.
The first method~\cite{nguyen2017identifying} verified the word distribution with Zipf's law. 
In the second method, ~\newcite{li2015machine} extracted features from the parsing tree and used them for classifiers.
The most similar method to our coherence features~\cite{nguyen2019detecting} matched similar words within the text and manipulates on maximum similarity.
Finally, the last method~\cite{aharoni2014automatic} combined POS $N$-gram model with functional words to identify the machine text.

The word-distribution-based method~\cite{nguyen2017identifying} is suitable for large text, e.g., document and web page, because of needing large words to adapt to the Zipf's law; but the performance is positively affected on the sentence level.
On the other hand, the syntax-based method~\cite{li2015machine} seems more appropriate with this task but this work merely focuses only on text structure and dismisses the intrinsic meaning.
Besides that, the previous coherence-based method~\cite{nguyen2019detecting} only used a maximum similarity of each word rather than near-optimal similarities.
Therefore, this work is more appropriate to a paragraph than a sentence, which has a limit number of words.
In another approach, the adding of function words into POS $N$-gram model~\cite{aharoni2014automatic} can improve the SMO(SVM) classifier.
Among these classifiers, our method is the most stable especially in both SVM classifiers with the highest accuracy of 82.0\%.

\section{Conclusion}
\label{section:conclusion}

We have investigated the issues from one of the most harmful adversarial texts which are generated by changing the structures of the original texts.
Although the adversarial generators can produce understandable texts, which preserve the meaning of the origins; the coherence and fluency of the generated texts still have limits.
Moreover, a person has a higher probability to create a professional text by using flexible words.
In another aspect, the optimization process leads to the adversarial texts incurring some artificial phenomenal such as a shortage in length or duplication in phrases. 
Based on the findings, we propose a method to identify the adversarial texts by suggesting distinguishable features with the original texts.
The results of the evaluation on the adversarial texts generated from the movie review corpus show that our proposed method achieves high performance: 82.0\% of the accuracy and 18.4\% of the EER which are greater than related methods with the best accuracy 77.0\% and EER 22.8\%.

In future work, we improve the proposed features by using deep learning networks and identify other harmful adversarial texts such as product reviews and political comments. 
We also improve the quality of useful machine-generated texts based on our analysis in this paper.

\bibliographystyle{acl}
\bibliography{adversarial_text_detection} 

\end{document}